\documentclass[letterpaper, 10 pt, conference]{ieeeconf}
\IEEEoverridecommandlockouts
\usepackage{amsmath}
\usepackage{graphicx}
\usepackage{psfrag}
\usepackage{amsmath}
\usepackage{amssymb}
\usepackage{graphicx}
\usepackage{subfigure}
\usepackage{verbatim}
\usepackage{latexsym}
\usepackage[dvipsnames]{xcolor}
\usepackage{cite}
\usepackage{stmaryrd}
\usepackage{pifont}
\usepackage{amssymb}
\usepackage{gensymb}
\usepackage{colortbl}
\usepackage{multirow}
\usepackage{caption}
\usepackage{hyperref}

\usepackage{epstopdf}
\usepackage{cleveref}
\usepackage{amsmath}
\usepackage{tabto}
\usepackage{cite}
\usepackage{tikz}
\usepackage[linesnumbered,ruled,vlined]{algorithm2e}

\graphicspath{./images}
\DeclareGraphicsExtensions{.pdf,.eps, .jpg, .png, .PNG}

\title{\LARGE \bf Development of a Steel Bridge Climbing Robot}
\vspace{-15pt}
\author{Son Thanh Nguyen, Hung Manh La, \textit{IEEE Senior Member}
\thanks{This work is supported   by the U.S. Department of Transportation, Office of the Assistant Secretary for Research and Technology (USDOT/OST-R) under Grant No. 69A3551747126 through INSPIRE University Transportation Center.
}
\thanks{The authors are with the Advanced Robotics and Automation (ARA) Lab, Department of Computer Science and Engineering, University of Nevada, Reno, NV  89557, USA. Corresponding author: Hung La, email: hla@unr.edu.}
}

\begin{document}
\vspace{-15pt}

\maketitle

\begin{abstract} 
Motivated by a high demand for automated inspection of civil infrastructure, this work presents a new design and development of a tank-like robot for structural health monitoring. Unlike most existing magnetic wheeled mobile robot designs, which may be suitable for climbing on flat steel surface, our proposed tank-like robot design uses reciprocating mechanism and roller-chains to make it capable of climbing on different structural shapes (e.g., cylinder, cube) with coated or non-coated steel surfaces. The proposed robot is able to transition from one surface to the other (e.g., from flat surface to curving surface).
 Taking into account of several strict considerations (including tight dimension, efficient adhesion and climbing flexibility) to adapt with various shapes of steel structures, a prototype tank-like robot
incorporating multiple sensors (hall-effects, sonars, inertial measurement unit and camera), has been developed. Rigorous analysis of robot kinematics, adhesion force, sliding failure and turn-over failure has been conducted to demonstrate the stability of the proposed design. Mechanical and magnetic force analysis together with sliding/turn-over failure investigation can serve as an useful framework for designing various steel climbing robots in the future. Experimental results and field deployments confirm the adhesion and climbing capability of the developed robot. 
\end{abstract}


\section{Introduction}

The need for automated inspection of civil infrastructures is growing since the current inspection practice is mainly manual and can not meet the demand for frequent and adequate inspection and maintenance \cite{USFHWA, MCCREA2002}. Civil infrastructures like bridges in general or steel bridges in particular is performed by inspectors with visual inspection or using chain dragging for crack and delamination  detection, which are very time consuming and not efficient. Often, it is dangerous for the inspectors to climb up and hang on cables to inspect high structures of bridges \cite{Golden_Gate_Inspection2018}. Additionally, some areas of the structures are hard to reach or may not be  accessible  due to their confined space. 


As an effort to automate the inspection process, there has been some implementations of climbing robots for inspection \cite{Fischer_IC2008, Fischer_TIE2011, Ratsamee_SSRR2016, Wang_TRO2017, La_IJFR2017}. A legged robot that can transition across structure members for steel bridge inspection was developed \cite{Mazumdar}. The robot uses permanent magnets integrated with each foot to allow it to hang from a steel bar.  In another case, a magnetic wheeled robot, which can  carry magneto resistive sensor array for detecting corrosion and crack, was developed \cite{RWang}. Similarly, several climbing permanent magnet-robots  \cite{Fabien_IJFR2009, Pack, Abderrahim, Leibbrandt, Rodriguez, SanMillan, Shen_MA20005, Tavakol_RAS2013, Eich_MED2015} were designed to carry non-destructive evaluation (NDE) devices to detect corrosion, weld defect and crack, and these robots can be applied for inspecting steel structures and bridges. Other efforts has shown for development of climbing robots for power plant inspection \cite{Gilles_IJFR2012}, bridge cable inspection \cite{Cho_TMECH2013} and  automated  bridge deck inspection  \cite{La1,La2,La3, Lim1, Lim2, Gibb_IJFR2018, Sutter_AC2018}. Additionally, significant development of climbing robots for steel structure and bridge inspection has been reported in \cite{Zhu_TMECH2012, Lee_RAS2013, Seo_TMECH2013, Wang_ICA2014, Guo_ICRA14,  Kamdar2015, Ward2015ClimbingRF, PL_Allerton2016, Wang_IJIR2016,  PL_ISARC2016, Takada_inventions2017, La_Robotica_2018}. 

In summary, most existing designs  do not own flexible configurations  and thus make them hard to adapt to different steel structures. Also, there is a lack of rigorous analysis of robot's kinematics, adhesion force, sliding failure and turn-over failure that would lead to inefficient design. More importantly, none of the above mentioned climbing designs have been deployed and validated on real steel bridges. 


This paper presents a new design and implementation of a climbing robot  to provide a practical solution for steel structure inspection (e.g., bridges, poles, pipes, etc.) The proposed small tank-like robot with reciprocating mechanism features various deformable 3D configurations, which can allow it to transition among steel structure members for efficient inspection. The robot utilizing adhesion force generated by permanent magnets is able to well adhere on steel structures while moving. The roller-chain design allows the robot to overcome obstacles including nuts, bolts, convex and concave conners. To demonstrate the robot's working principle, it has been deployed for climbing on more than 20 steel bridges.



\section{Overall Design}
\label{sec:OverallDesign}

\begin{figure}[!htbp]
\centerline{\includegraphics[width=\linewidth]{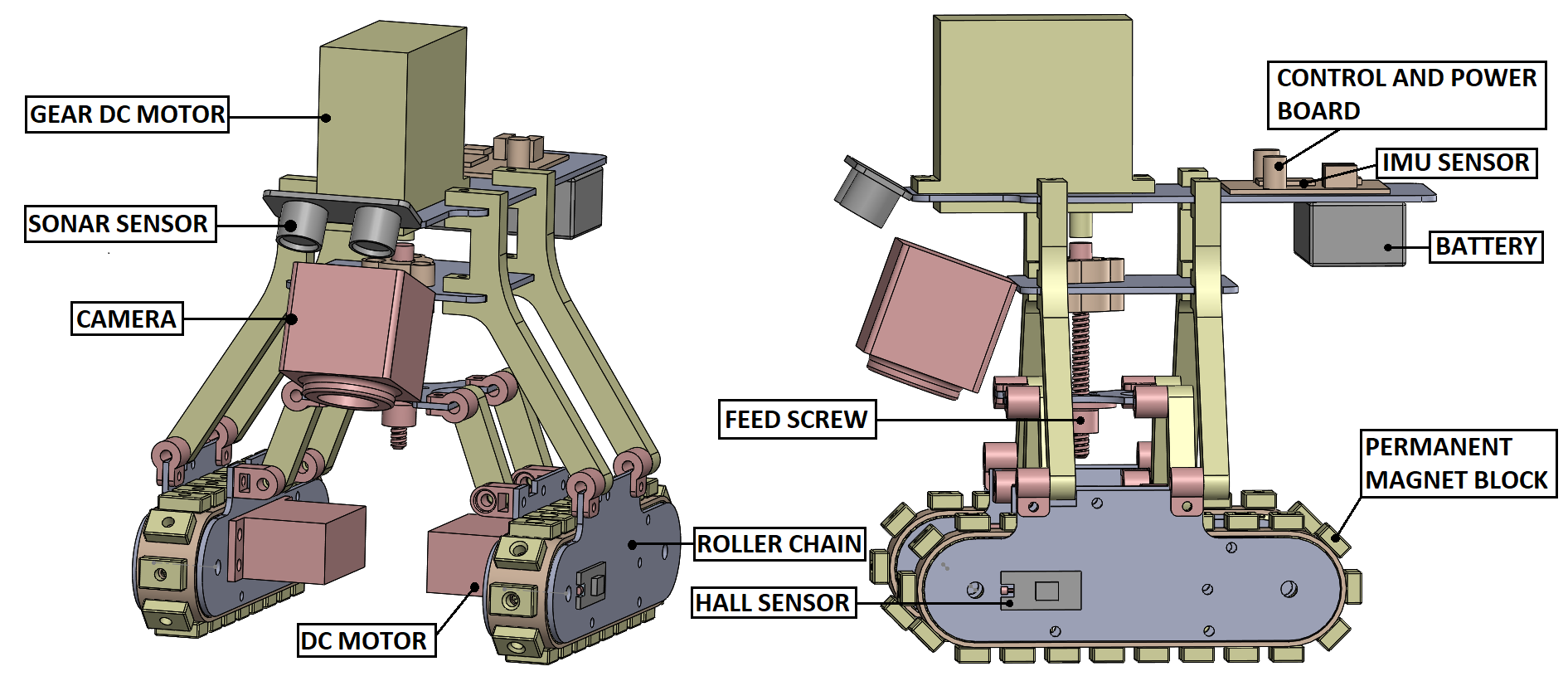}}
\caption{Steel climbing robotic system.}
\label{fig:FullRobot}
\vspace{-10pt}
\end{figure}

The overall design concept of the climbing robot is shown in Fig. \ref{fig:FullRobot}. 
The roller-chains embedded with permanent magnets for adhesion force creation enable the robot to adhere to steel surfaces without consuming any power. The control architecture of the robot consists of a low-level and a high-level controllers. The low-level controller handles low-level tasks including (i) converting velocity and heading command from the high-level controller to Pulse Width Modulation (PWM) data to drive motors, and (ii) reading data from multiple sensors for navigation purposes. The high-level controller is embedded in an onboard computer to enable data processing and ground station communication. Both controllers fuse sensor data to provide desired linear velocity and heading for the robot and acquire data from advanced sensors. Furthermore, the high-level controller sends data wirelessly to ground station  for  processing and logging.

The robot is equipped with various sensors for navigation as well as steel structure evaluation. There is a video camera for visual image capturing and video streaming.   The robot has two roller-chains, and  each roller-chain is integrated with two hall-effect sensors, which are mounted next to each other and close to robot's roller. 
Since the magnet block inside each roller-chain will move when the robot moves, we can extract the velocity and traveling distance of each roller-chain after combining the data from these two hall-effect sensors. Additionally, an Inertial Measurement Unit (IMU) is used for the robot's localization. 
Besides, to avoid falling off, the robot has sonar sensors mounted at front of the robot to detect if a surface underneath exists.


\section{Mechanical Design and Analysis}
\label{sec:MecDesign}

A tank-like robot mechanism  design is proposed to take advantage of the flexibility in maneuvering. Two motors are used to drive two roller-chains, and another  motor is used to drive the transformation of robot to approach different contour surfaces. The robot's parameters are shown in Table \ref{tab:robotparams} while motor's parameters are listed in Table \ref{tab:motorparams}.
\begin{table}[ht]
\begin{center}
  \caption{Robot Parameters.}
  \begin{tabular}{|c|c|}
     \hline
\textbf{Length} & 163 mm \\
     \hline
\textbf{Width} & 145 mm \\
     \hline
\textbf{Height} & 198 mm \\
     \hline
\textbf{Weight} &  3 kg  \\
     \hline
\textbf{Drive} & 2 motorized roller-chains  and 1 motorized transformation\\
     \hline
  \end{tabular}
  \label{tab:robotparams}
\end{center}
\vspace{-15pt}
\end{table}

\begin{table}[ht]
\begin{center}
  \caption{Motor Parameters.}
  \begin{tabular}{|c|c|c|}
     \hline
&\textbf{Moving motors} & \textbf{Tranforming motor}\\
     \hline
\textbf{Torque} & 12 kg.cm (2S Li-Po)& 32 kg.cm\\
     \hline
\textbf{Speed} & 0.12 sec/ ${60}^{\circ}$ (2S Li-Po)& 0.15 sec/ ${60}^{\circ}$ (2S Li-Po)\\
     \hline
\textbf{Length} & 40.13 mm& 60.5 mm \\
     \hline
\textbf{Width} & 20.83 mm& 30.4 mm\\
     \hline
\textbf{Height} & 39.62 mm& 45.6 mm\\
     \hline
\textbf{Weight} & 71 g& 156 g\\
     \hline
\textbf{Voltage} & 6-8.5V (2S Li-Po battery)&6-8.5V (2S Li-Po battery) \\
     \hline
  \end{tabular}
  \label{tab:motorparams}
\end{center}
\vspace{-10pt}
\end{table}

\begin{figure}[!htbp]
\centerline{\includegraphics[width=\linewidth]{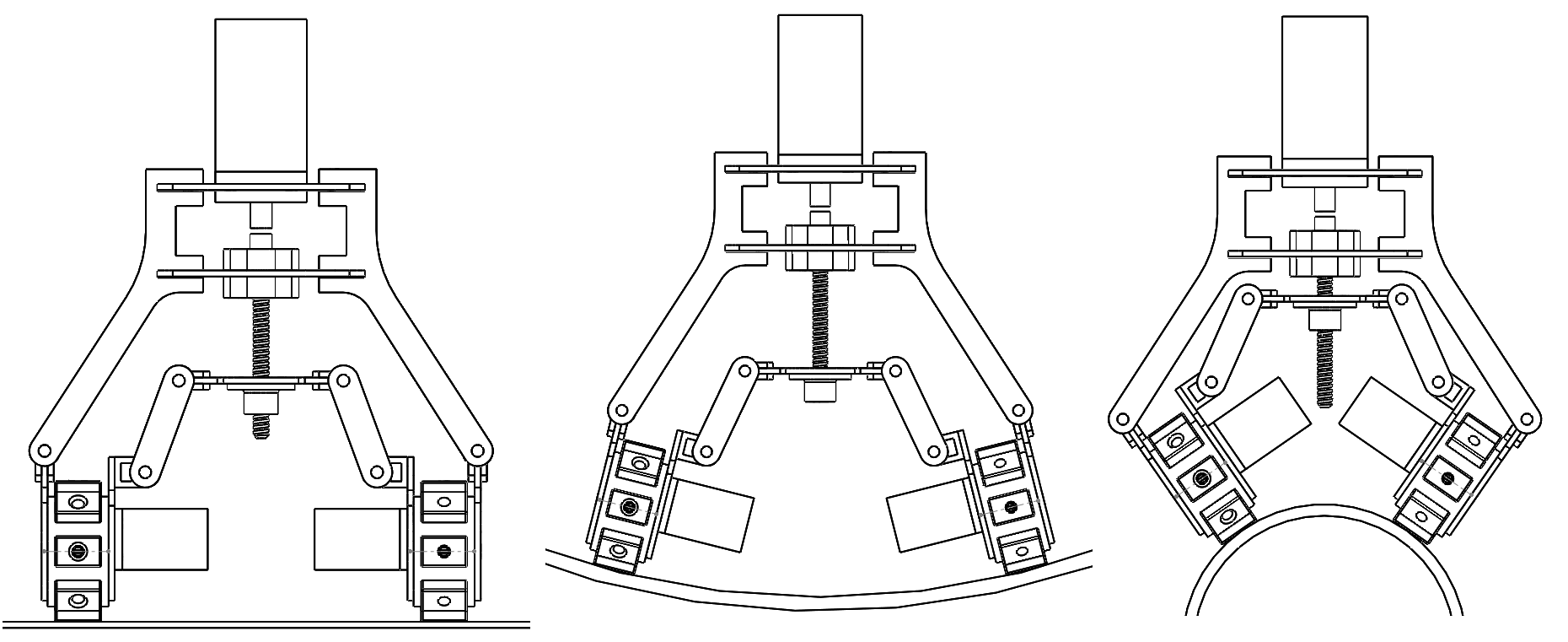}}
    \caption{Reciprocating mechanism for robot transformation.}
    \label{fig:mechanism}
    \vspace{-10pt}
\end{figure}
\begin{figure}[ht]
\centerline{\includegraphics[width=\linewidth]{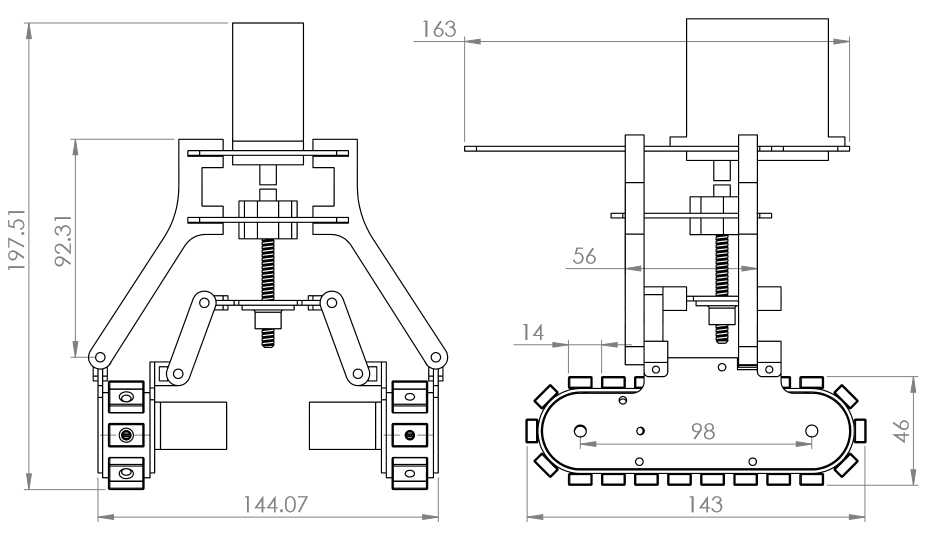}}
    \caption{Specifications of the robot's design (unit in millimeter).}
    \label{dimension}
\vspace{-10pt}
\end{figure}

A roller-chain is designed to carry 22 Neodymium magnet blocks with poles on flat ends as shown in Fig. \ref{fig:FullRobot}. At each motion, there will be maximum of 8 magnet blocks contacting the flat steel surface. 
Roller-chains are designed to enable the robot to overcome several real climbing scenarios including transitioning among surfaces with different inclination levels ($0-90^{\degree}$ change in orientation) or getting rid of being stuck. Reciprocating mechanism has been added in order to transform the robot to adapt with  different contour surfaces as shown in Fig. \ref{fig:mechanism}. Specifications of the robot's design is shown in Fig. \ref{dimension}.

As aforementioned, at each moment there are maximum 8 magnet blocks in the chain contacting the steel surface. Hence, we have  the magnetic force created by each robot's roller-chain as: $\sum{F}_{{m}_{j}}(j=1:8)$.  Since the robot has two roller-chains, the total magnetic adhesion force, $F_{m}$, created by these two roller-chains is:
\begin{equation} \label{eq:magcal}
	F_{m} = 2 *\sum{F}_{{m}_{j}}(j=1:8).
\end{equation}

\subsection{Robot kinematics analysis   }
\label{sec:kinem atics}

\begin{figure}[htb]
\centerline{\includegraphics[width=0.85\linewidth]{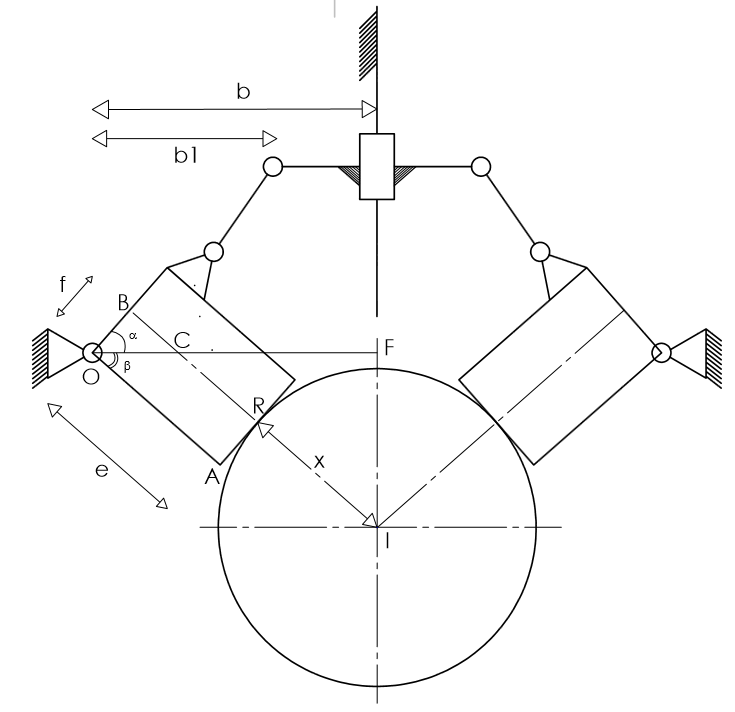}}
    \caption{Kinematics of the robot: $b = 72mm$, $b_{1} = 45mm$, $f = 11mm$, and $e = 55mm$. }
    \label{fig:Ki}
    \vspace{-15pt}
\end{figure}
\begin{figure}[htb]
\centerline{\includegraphics[width=0.8\linewidth]{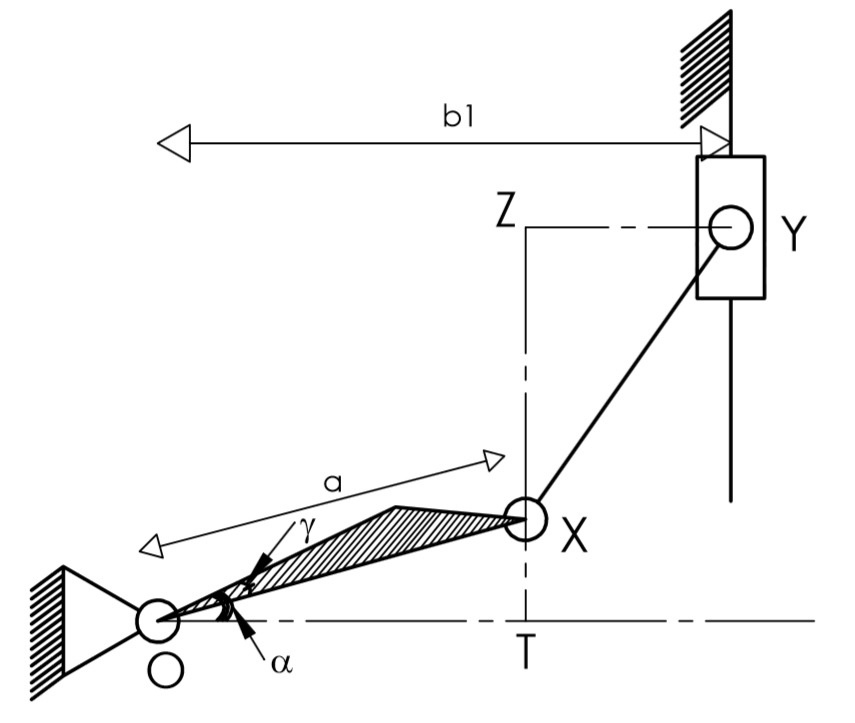}}
    \caption{Reciprocating mechanism: $a = 33.7mm$, $XY = 32mm$, and $\gamma = 12\degree$}
    \label{fig:Crank}
    \vspace{-15pt}
\end{figure}

Kinematics analysis of reciprocating mechanism is to calculate radius of steel cylinder $(x)$, that robot can climb on. The feed screw mechanism has gear ratio is $1:19$ with screw pitch is $0.8mm$.
 Fig. \ref{fig:Ki} presents kinematics of the robot, and Fig. \ref{fig:Crank} illustrates a general architecture of the robot's reciprocating mechanism. 

From Fig. \ref{fig:Ki}: $x=IC-CR$, in which, 
$IC=\dfrac {CF}{\cos{\beta}} =\dfrac{OF-OC}{\cos{\beta}}= \dfrac{b-\dfrac{f}{\cos{\alpha}}}{\cos{\beta}} = \dfrac {b}{\cos{\beta}} - \dfrac {f}{\cos{\alpha}\cos{\beta}}$;\\

$CR = e - BC = e - f \tan{\alpha}$
with  $\alpha + \beta = 90^{\degree}$;\\
\begin{equation} \label{k1} 
\rightarrow
x = \dfrac {b}{\cos{\beta}} - \dfrac {b}{\cos{\alpha} \cos{\beta}} - e + f\tan{\alpha}.
\end{equation}

From Fig. \ref{fig:Crank}: $y = XZ = \sqrt{{XY}^{2} -{YZ}^{2}}$, where
$YZ = b_1 - OT \Leftrightarrow YZ = b_1 - OX\cos{\phi}; (\phi = \alpha - \gamma)$;
\begin{equation}\label{k2} 
\rightarrow
y = \sqrt{{XY}^2 - ({b_1 - a\cos{\phi}})^2}.
\end{equation}

\begin{figure}[ht]
\centerline{\includegraphics[width=\linewidth]{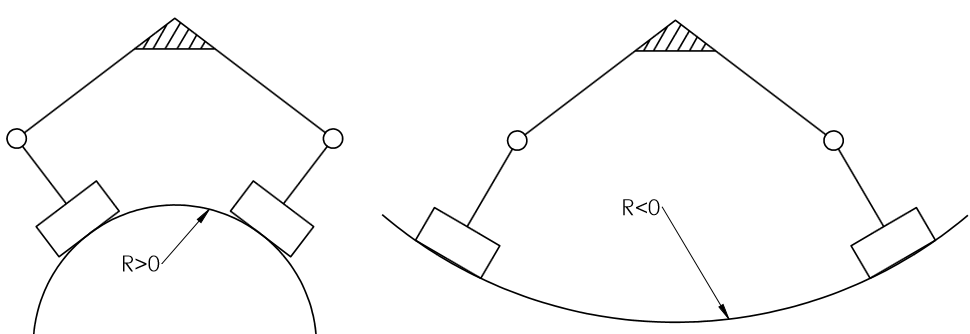}}
    \caption{Different cases when the robot moves on curving steel surface:\\Left-figure) Robot moves on positive curving surface;\\Right-figure) Robot moves on negative curving surface.}
    \label{fig:curve}
   \vspace{-10pt} 
\end{figure}

From equations (2), (3) and  the designed gear ratio (1:19), we can calculate radius $x$ of the steel cylinder based on rotations of the transformation motor. Robot is designed to work on steel cylinders having a smallest radius from $+5cm$ to $- 25cm$ as descriptions on Fig. \ref{fig:curve} with $7.5 cm$ feed screw movement. 

\begin{figure}[ht]
\centerline{\includegraphics[width=0.6\linewidth]{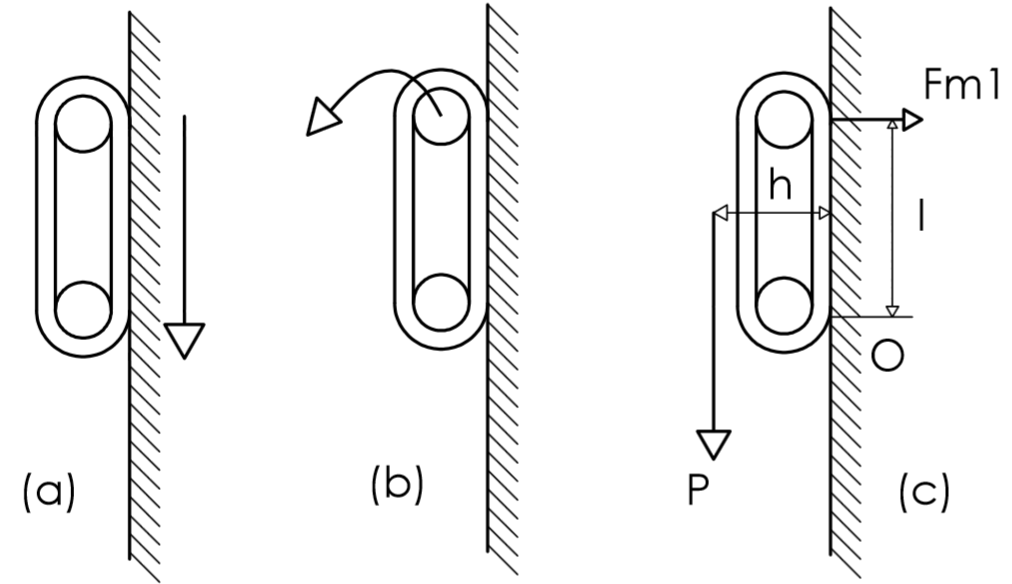}}
    \caption{a) Sliding failure; b) Turn-over failure; (c) Moment calculation at point $O$.}
    \label{fig:failure}
\vspace{-10pt}
\end{figure}

To maintain stability of the robot while climbing on steel structures, the sliding and turn-over failures as illustrated in Fig. \ref{fig:failure} (a,b) should be investigated.
\subsection{Sliding Failure Investigation}

\begin{figure}[ht]
\centerline{\includegraphics[width=\linewidth]{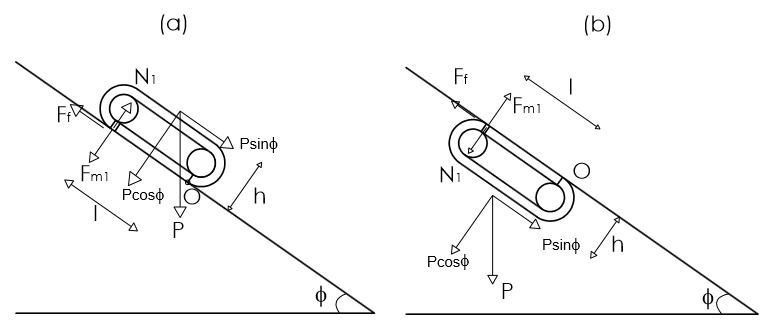}}
    \caption{a) Climbing on top inclined surface; b) Climbing underneath inclined surface.}
    \label{fig:sliding}
\vspace{-10pt}
\end{figure}

In general case, based on the proposed design, the robot is able to climb on different shapes of structures (cylinder, cube or flat) with different inclination levels as shown in Fig. \ref{fig:sliding}. Let $P$ be the robot's total weight ($P=mg$, where $m$ is the robot's mass, and $g$ is the gravitational acceleration). Let $\vec{F}_{m}$ be the magnetic adhesion force, $N$ be the reaction force, $\mu$ be the frictional coefficient, $F_f$ be frictional force, and $\alpha$ be the degree of inclination. Denote $\sum \vec{F}$ as the total force applied to the robot. Based on Newton's second law of motion, $\sum \vec{F} = \vec{0}$ when the robot stops.

When the robot climbs on top of an inclined surface (Fig. \ref{fig:sliding}(a)), based on our previous work \cite{PL_Allerton2016}, we can obtain  the magnetic adhesion force: $F_{m}=\dfrac{P\sin\phi}{\mu} -{P\cos\phi}.$ Hence, the sliding failure can be avoided if the magnetic force satisfies the following condition:
\begin{equation} \label{e9} 
	{F}_{m} > \dfrac{P\sin\phi}{\mu} - {P\cos\phi}.
\vspace{-5pt}
\end{equation}

When the robot climbs underneath an inclined surface (Fig. \ref{fig:sliding}(b)), we obtain:
$F_{m}=\dfrac{P\sin\phi}{\mu} + {P\cos\phi}.$ In this case, the magnetic force should be
\begin{equation} \label{e11} 
	{F}_{m} > \dfrac{P\sin\phi}{\mu} + {P\cos\phi}.
\vspace{-5pt}
\end{equation}

When the robot climbs on a vertical surface ($\phi={90}^{o}$)
\begin{equation} \label{e12} 
	{F}_{m} > \dfrac{P}{\mu} .
\vspace{-5pt}
\end{equation}

From equations (\ref{e9}), (\ref{e11}) and (\ref{e12}), to avoid sliding failure in any cases, the magnetic force should be
\begin{equation} \label{e13} 
	{F}_{m} > max\Big\{ \dfrac{P\sin\phi}{\mu} - {P\cos\phi} ;\dfrac{P\sin\phi}{\mu} + {P\cos\phi}\Big\}.
\vspace{-5pt}
\end{equation}

Since $0 < \phi\leq 90 \Rightarrow \cos\phi\ge 0$, the overall condition for avoiding sliding failure is
\begin{equation} \label{e15} 
	{F}_{m} > \dfrac{P\sin\phi}{\mu} + {P\cos\phi}.
\vspace{-5pt}
\end{equation}
Assume that the frictional coefficient $\mu$ between two roller-chains and steel surface is from $[0.4 - 0.8]$, we see that \Big($\dfrac{\sin\phi}{\mu} + \cos\phi$\Big) decreases when $\mu$ increases, or we have:
\begin{gather*}
	0.4 \leq \mu \leq 0.8; 0 < \phi \leq 90 \\
	\Rightarrow max\big\{ \dfrac{\sin\phi}{\mu} + \cos\phi \Big\} = 2.5 \Rightarrow {F}_{m} \ge {2.5P}.
\end{gather*}
In summary, the robot's magnetic adhesion force  should be greater or equal to $2.5$ of the robot's weight. 

\subsection{Turn-over Failure Investigation}
Let $l$ be the distance between first and last magnet block contacting to the surface, and $h$ be the distance between the center of mass to the surface (Fig. \ref{fig:failure}(c)). Moment at point $O$  (the point that the first magnet block contacts the steel surface) is calculated as follows:
\begin{align*} \label{e17}
	\sum M &= {P} * h- 2F_{m_1}*l = 0
	\Leftrightarrow {F}_{m_1}=\dfrac{P*h}{2l}.
\vspace{-5pt}
\end{align*}
To avoid turn-over failure, the magnetic force of the first contacting magnet block:
\begin{equation} \label{e18} 
	{F}_{m_1}>\dfrac{P*h}{2l}.
\vspace{-5pt}
\end{equation}
From (\ref{e18}), to avoid the failure we can lower $\dfrac{h}{l}$,  which means making the robot's center of mass closer to the steel surface. In the proposed design (Fig. \ref{dimension}), $h=4.6$cm, and the total robot height $h_{r}= h+ 9.231cm = 19.751$cm. Therefore, to avoid both sliding and turn-over failures, the robot's magnetic force of each magnet block should satisfy:
\begin{equation} \label{e19} 
	{F}_{m_j}(j=1:n) > max\Big\{\dfrac{2.5P}{n};\dfrac{P*h_{r}}{2l}\Big\}.
\end{equation}
Following the proposed design,  $P=3$kg, $n=16$ magnet blocks (each roller-chain has maximum 8 magnet blocks contacting the steel surface), $l=9.8$cm, or $F_{m_j} (j=1:16)> max\Big\{\dfrac{2.5 * 3}{16};\dfrac{3*19.751}{2*9.8}\Big\} > 3(N)$. We conducted some tests as discussed in subsection \ref{sec:Pullforce} to make sure the proposed design satisfied this condition.

\subsection{Motor Torque Analysis}

Apart from the magnetic force analysis, we have conducted another analysis to determine the appropriate motor to drive the robot. In order to make the robot move, the force created by the motor should win the adhesion force of the last permanent magnet and the steel surface. As shown in Fig. \ref{fig:curveAnalysis1}, denote $M$ as the torque of one motor, $Q$ is rotation fulcrum, $i$ is the arm from $F_{m_j}$ to $Q$. Assume that the total driving force of the robot is the sum of two motor's forces, the required moment is satisfied:
\begin{figure}[ht]
\centerline{\includegraphics[width=\linewidth]{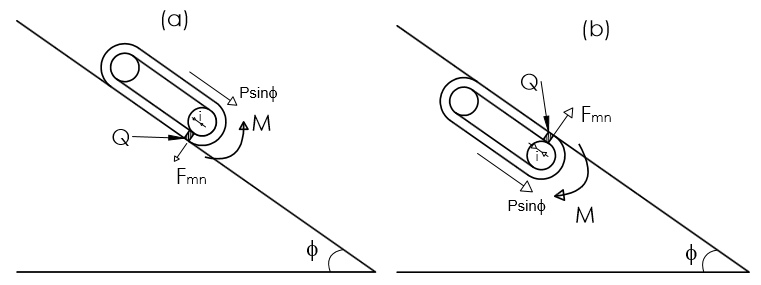}}
   \caption{a) Robot moves on top of inclined surface; b) Robot moves on bottom of inclined surface.}
    \label{fig:curveAnalysis1}
    \vspace{-15pt}
\end{figure}\\
\begin{equation*}
M > h_r*\dfrac{P\sin\phi}{2} + F_{m_j}(j=1:n)*i/g.
\end{equation*}

When the robot moves on a vertical surface, maximum $\phi=90^0$
\begin{equation} \label{e29}
M > h_r*\dfrac{P}{2} + F_{m_j}(j=1:n)*i /g. 
\end{equation}
$\rightarrow M> 31.65 (kg.cm)$ with $h_r=19.751cm, P=3kg, g = 9.8, F_{m_n}=39.7 N$ (getting this number from the magnet's data sheet), $i = 5mm = 0.5cm$.
The selected moving motor has $12kg.cm$ moment and gear ratio is $11:20$, hence the total moment of the robot with two motors is $43.64 kg.cm$, which satisfies  (\ref{e29}).

\begin{figure}[ht]
\centerline{\includegraphics[width=\linewidth]{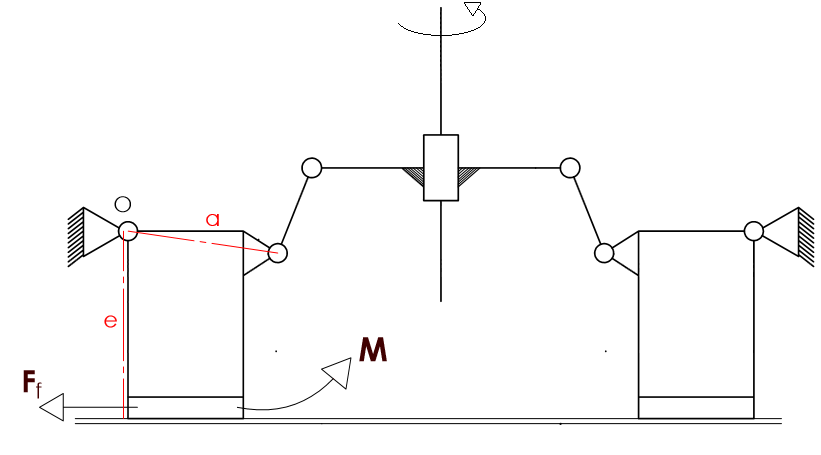}}
    \caption{Transformation mechanism with 7 joints to create climbing flexibility of the robot.}
    \label{fig:tranformation}
\vspace{-5pt}
\end{figure}

Regarding transformation mechanism as shown in Fig. \ref{fig:tranformation}, when the robot moves from flat surface to curving surface or reversing, the transforming mechanism works to make sure that the roller-chains contact  steel surfaces with best condition. 
The transformation motor through mechanical system creates moment $(M)$ to release magnet blocks from steel surface (see Fig. \ref{fig:tranformation}). However, the calculation gear ratio of reciprocating mechanism is not simple due to its nonlinear, as showed in (\ref{k2}). We choose the special case when friction is strongest for calculation moment of transforming motor. The moment has to satisfy:

$M > F_f * e \Leftrightarrow M> e* (k.N) \Leftrightarrow M>e * k (P+F_m)$.\\
\begin{equation} \label{e30}
\Rightarrow M>e * k (P+nF_{m_j}(j=1:n)).
\end{equation}

From equation (\ref{k2}) and kinematics parameter of reciprocating mechanism, it is straight forward to calculate the gear ratio in this case. The ratio $\alpha$ : $y$ is approximately $1^{\degree}: 1 mm$. Besides, the ratio of $a$ : $e = 1 : 1.7$, and the gear ration of the feed screw is $0.8 mm$ : $360^{\degree}$. As result, a total gear ratio of whole system is $26.5 : 1$. 
From equation (\ref{e30}) and assumption that system efficiency is 80\%, the required moment of motor is $> 13,75 kg.cm$. The selected transformation motor with $32 kg.cm$ is satisfied.

\section{Robot Control}

\begin{figure}[ht]
\centerline{\includegraphics[width=0.7\linewidth]{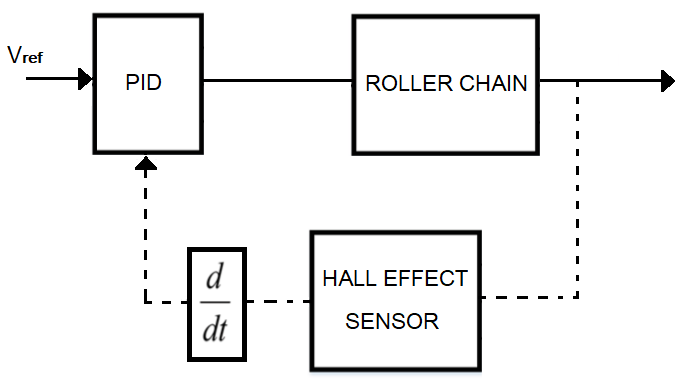}}
    \caption{Single roller's velocity control.}
    \label{fig:PIDvel}
    \vspace{-10pt}
\end{figure}
\begin{figure}[ht]
\centerline{\includegraphics[width=\linewidth]{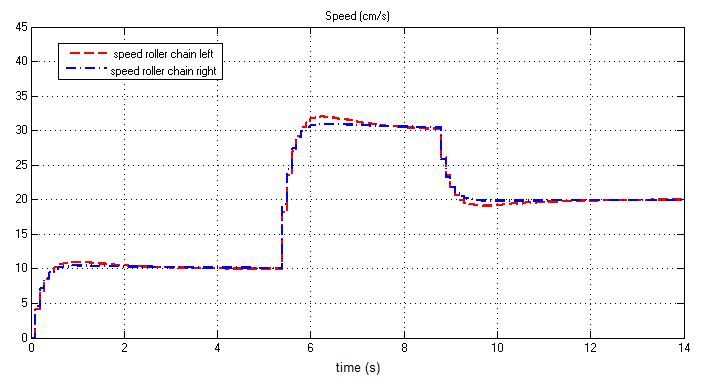}}
    \caption{Response of roller chains speed control.}
    \label{fig:responsePI}
    \vspace{-10pt}
\end{figure}

The speeds of the left and right roller-chains are not always equal and stable due to environmental noise or our imperfect model. The PID controller is applied to synchronize the speed of two roller-chains as show on Fig. \ref{fig:PIDvel}. Maximum speed recorded is around 35 cm/s, which is suitable for investigation task. The speed responses of left and right roller-chains are illustrated on Fig. \ref{fig:responsePI} with three reference speeds 10, 20 and 30 cm/s with the sample time  of 0.1s. We can see that the speeds of left and right roller-chains are well synchronized and follow the reference speeds very well.

When the robot travels on curving surfaces, reciprocating mechanism is driven automatically by driving the motor based on IR sensor signals. A PID controller is also applied to maintain the distance between the IR sensor and the surface. This is to keep approaching area of permanent magnet with steel surface consistent. Due to limited space, detail of this PID design is omitted.




\section{Robot Deployment}
\label{sec:Result}

To evaluate design and performance of the robot, experiments  for evaluating the magnetic force created by roller-chains have been conducted. The ability of climbing and failure avoidance were tested. During the test, a Lipo battery (2 cells) 7.4V 900 milliampere-hour $(mAh)$ is used to power the robot for about 1 hour  of working. One laptop which can connect to a wireless LAN is used as a ground station. The robot's mass $m=3kg$, and if we assume that the gravitational acceleration $g=10m/s^2$, the total weight of the robot is approximately $P=mg=30N$. Since $15mm\times 10mm\times 5mm$ magnet blocks are used, the total magnetic force is calculated as
\begin{equation} \label{eq:calculatedforce}
	\left\{\begin{array}{ll}
		F_{m_{j}}(j=1:8) = 39.7 (N)\\
		F_{m} = 2*8*F_{m_j} = 635.2 (N)
	\end{array}\right.
\end{equation}
which satisfies magnetic force condition as presented in equation (\ref{e19}). 

\subsection{Adhesion Force Measurement}
\label{sec:Pullforce}

\begin{figure}
\centerline{\includegraphics[width=0.8\linewidth, height = 4cm]{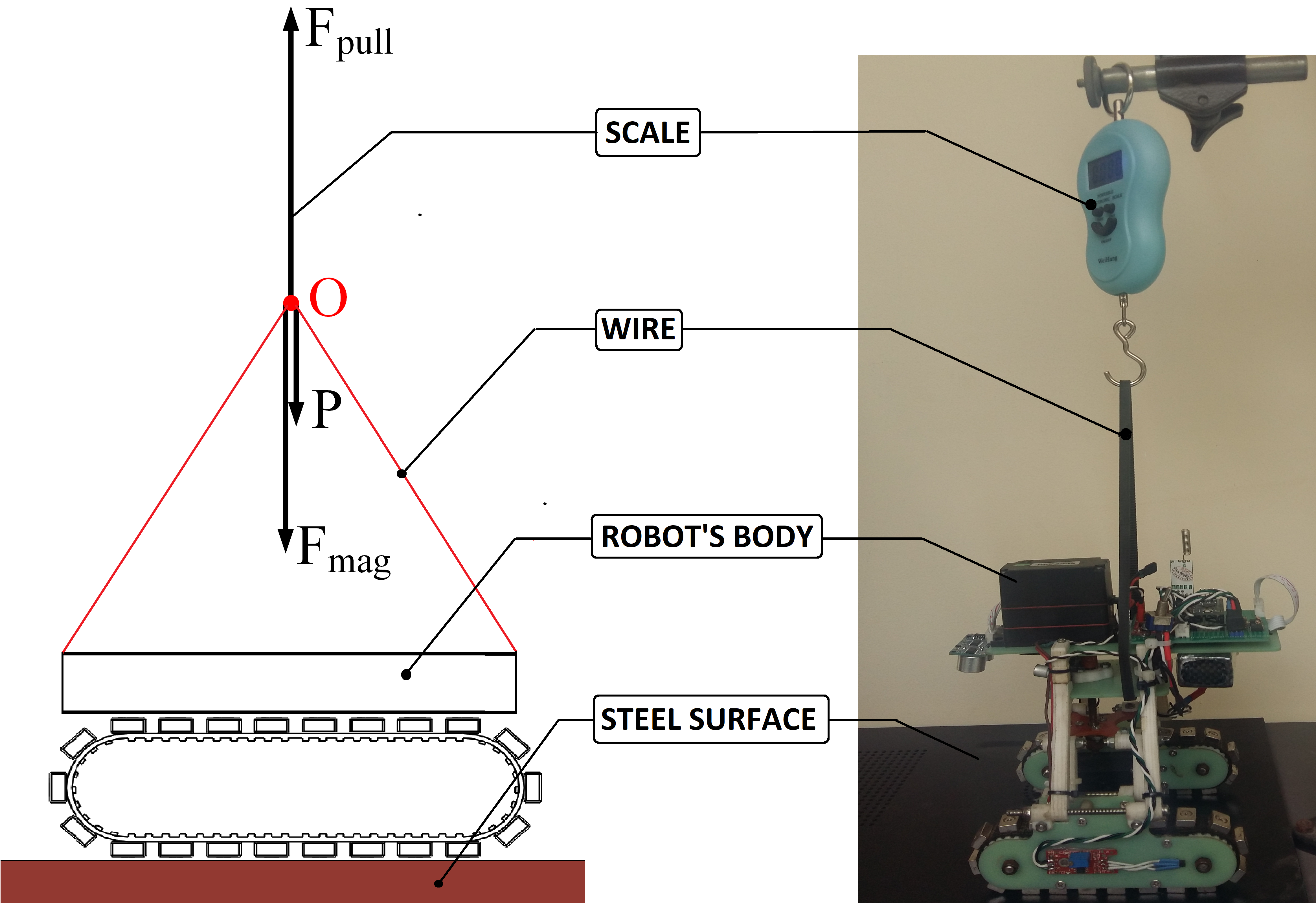}}
    \caption{Experimental setup for magnetic force measurement.}
    \label{fig:pullforcemeasure}
\vspace{-15pt}
\end{figure}
In order to measure the adhesion force created by permanent magnet, we have setup an environment as shown in Fig. \ref{fig:pullforcemeasure}. The robot's body - whose mass is $m=3kg$ - is placed on top of a flat steel surface while it is connected to a scale through an inelastic wire. We create a pull force onto the scale trying to lift the robot off the surface. At the time the robot is about to be off the surface, the force applied to the scale is equal to the sum of robot's weight and the magnetic pull force. Denote $F_{pull}$ as the force we applied onto the scale, $M$ is the value shown on the scale while $P$ is the weight of robot's body, and $F_{mag}$ is the magnetic force. With $g=10m/s^2$, $P=mg=30N$ and $F_{pull}=Mg=10M$, we can calculate magnetic adhesion force as follows
\begin{equation*}
F_{pull} = P + F_{m} \Rightarrow F_{m} = F_{pull} - P
\end{equation*}
\begin{equation} \label{eq:forcemeasure} 
	\Rightarrow F_{mag} =10M-3(N).
\vspace{-10pt}
\end{equation}

\begin{figure}[htb]
\centerline{\includegraphics[width=0.8\linewidth]{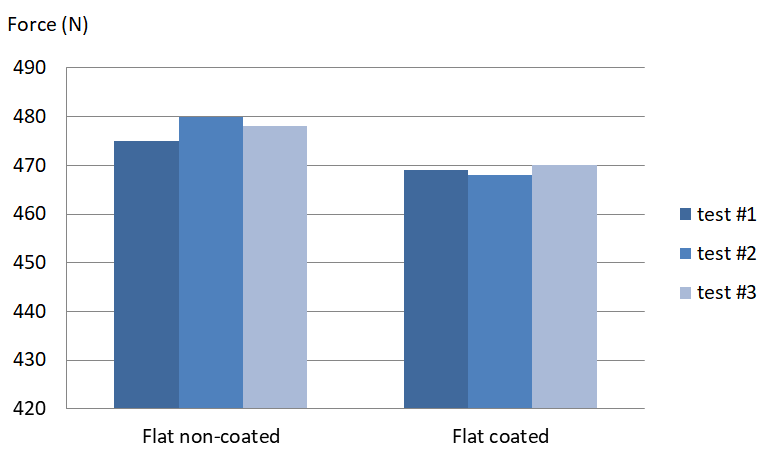}}
    \caption{Magnetic force measurements on coated and non-coated flat steel surface.}
    \label{fig:pullforcetest1}
\vspace{-10pt}
\end{figure}
\begin{figure}[htb]
\centerline{\includegraphics[width=0.8\linewidth]{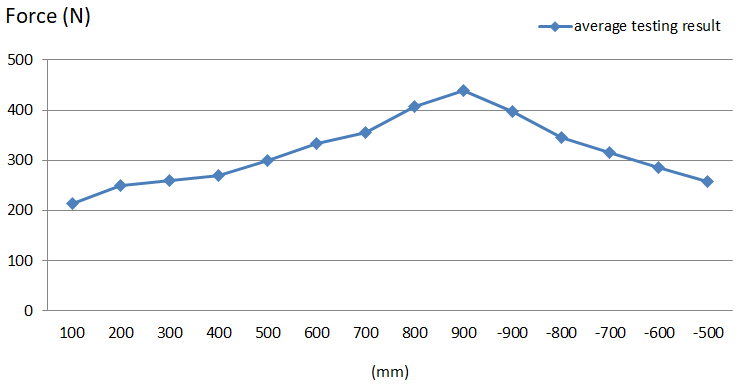}}
    \caption{Magnetic force measurements on curving steel surface with diameters (D) ranging from $100mm$ to $900mm$.}
    \label{fig:pullforcetest2}
\vspace{-10pt}
\end{figure}

Multiple tests have been conducted to measure the pull force when the robot's body is placed on different surfaces. The first test is on a flat non-coated steel surface, the second one is on a flat coated steel surface while the third test is on curve coated steel surfaces (positive and negative sides) with diameters (D) ranging from $100mm$ to $900mm$. All the tests are executed three times, and the results are presented in Fig. \ref{fig:pullforcetest1} and \ref{fig:pullforcetest2}.

According to equation (\ref{e19}), the magnetic force for each magnet block should be $F_{m_j} (j=1:n)> 3 (N)$ to avoid sliding and turn-over failure. Since at each time the robot has 16 magnet blocks which physically contact the steel surface, the total required adhesion force should be greater than $3 \times 16 = 48 (N)$. We can see that in Fig. \ref{fig:pullforcetest1} and \ref{fig:pullforcetest2} the minimum magnetic force is $210 (N)$, which is much greater than the required one, $48 (N)$. Therefore the robot adheres well on both coated/non-coated flat and curving surfaces. 



\subsection{Climbing Validation}

\begin{figure}[ht]
\centerline{\includegraphics[width=\linewidth]{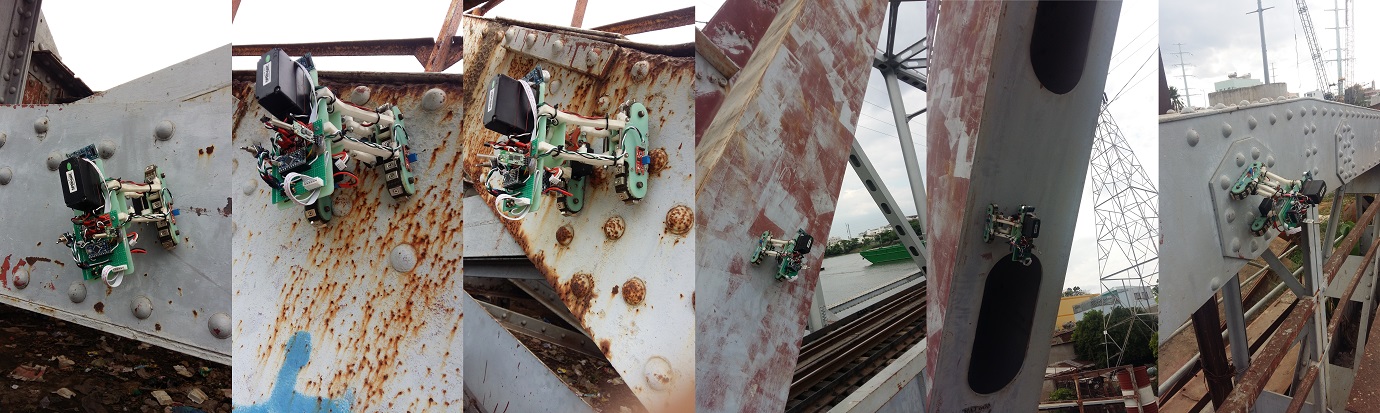}}
    \caption{Adhesion and climbing test on a rusty paint-coated steel bridge: flat structure with bolts/nuts. Video click here: \url{https://ara.cse.unr.edu/?page_id=11}}
    \label{fig:a4}
\vspace{-10pt}
\end{figure}

The outdoor experiments and robot deployments are conducted on more than 20 steel bridges. Due to limited space, only two typical climbing examples are shown in Fig. \ref{fig:a2}-\ref{fig:a4}. The steel structures have different thicknesses of paint coated on steel surfaces. Some paint-coated steel surfaces are very rusty, some are not clean, and some others are still in fine condition (minor-rusty). 

The robot is able to adhere very well on these steel structures while climbing. Even for the case the steel surface is curving (Fig. \ref{fig:a2}), the robot can still adhere tightly to the steel structures while performing the climb.  For rusty steel surfaces, it also shows strong climbing capability (Fig. \ref{fig:a4}). For more details please see the submitted video and this link:
\url{https://ara.cse.unr.edu/?page_id=11}


\begin{figure}[ht]
\centerline{\includegraphics[width=\linewidth]{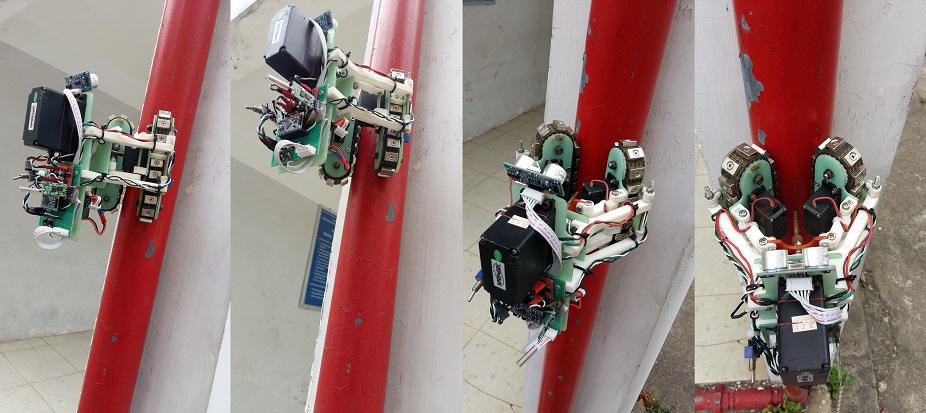}}
    \caption{Adhesion and climbing test on a thin paint-coated steel structure: cylinder shape D=100mm.}
    \label{fig:a2}
\vspace{-10pt}
\end{figure}


\begin{figure}[ht]
\centerline{\includegraphics[width=\linewidth, height=7cm]{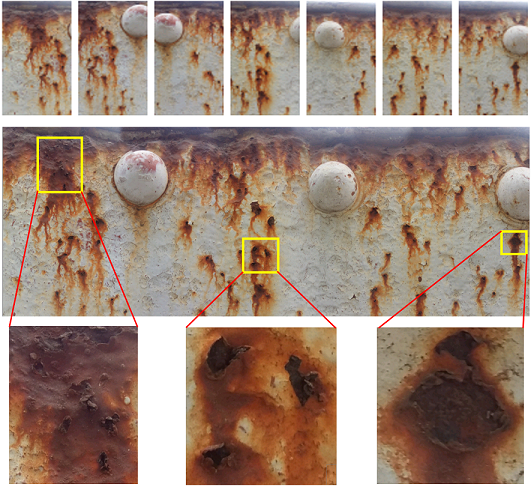}}%
\caption{Images stitching result: (Top) 7 individual images taken by the robot climbing on a bridge in Fig. \ref{fig:a4}; (Middle) Stitching image result from those 7 individual images; (Bottom) Closer look (zoom-in) at some areas, which has serious rusty condition with holes on the surface.}
\label{fig:stitch}
\vspace{-15pt}
\end{figure}

The robot is controlled to move and stop at every certain distance (e.g. 12cm ) to capture images of steel surface and send to the ground station. To enhance steel surface inspection, acquired images are then stitched together to produce an overall image of steel surface as shown in Fig. \ref{fig:stitch}. The image stitching is followed by our previously developed algorithm \cite{La2}.

\section{Conclusion and Future Work}
\label{sec:Conclusion}

This paper presents a new  development of a small tank-like robot, which is capable of climbing on different steel structure shapes to perform inspection. The robot design is implemented and validated on climbing on more than 20 different steel bridges. During the tests, the robot is able to firmly adhere on steel structures with various inclination levels. Rigorous analysis of magnetic adhesion force  has been performed to confirm that the robot is able to adhere to both flat and curving steel surfaces in various conditions (coated, non-coated and/or rusty.)  Various experiments have been conducted including magnetic force measurement, indoor and outdoor climbing tests in order to validate the force analysis as well as the climbing capability of the robot on different steel surfaces. The results show that when the magnetic adhesion force requirement is met, the robot is able to move and transition safely between steel surfaces without any failures. Multiple sensors are integrated to assist the robot's navigation as well as data collection. Inspection data is collected and transferred to ground station for visualization and processing.  In conclusion, the key contribution of the paper is the novel design of a climbing robot, which can be used for steel bridge or steel structure inspection. The rigorous magnetic force analysis can serve as a framework to calculate and design different types of steel inspection robots in the future.

Further works needed to be done include localization using odometry, IMU and visual data; and implementation of map construction method as well as visual crack detection algorithm. The robot can be further equipped   with NDE sensors (e.g., eddy current, thermal sensors, etc.) for more in-depth inspection of steel structures. 


\bibliographystyle{unsrt}
\bibliography{RefFile}

\begin{thebibliography}{10}

\bibitem{USFHWA}
U.S~Department of~Transportation Highway~Administration.
\newblock National bridge inventory data.
\newblock {\em http://www.fhwa.dot.gov/bridge/nbi.cfm}.

\bibitem{MCCREA2002}
A.~McCrea, D.~Chamberlain, and R.~Navon.
\newblock Automated inspection and restoration of steel bridges -- a critical
  review of methods and enabling technologies.
\newblock {\em Automation in Construction}, 11(4):351 -- 373, 2002.

\bibitem{Golden_Gate_Inspection2018}
Crews inspect condition of golden gate bridge's towers, \textsc{A}pril 30,
  2018.
\newblock
  \url{https://www.nbcbayarea.com/on-air/as-seen-on/Crews-Inspect-Condition-of-Golden-Gate-Bridge_s-Towers_Bay-Area-481315951.html}.
\newblock Accessed: 2018-09-13.

\bibitem{Fischer_IC2008}
W.~Fischer, F.~T{\^a}che, and R.~Siegwart.
\newblock {\em Magnetic Wall Climbing Robot for Thin Surfaces with Specific
  Obstacles}, pages 551--561.
\newblock Springer Berlin Heidelberg, Berlin, Heidelberg, 2008.

\bibitem{Fischer_TIE2011}
W.~Fischer, G.~Caprari, R.~Siegwart, and R.~Moser.
\newblock Locomotion system for a mobile robot on magnetic wheels with both
  axial and circumferential mobility and with only an 8-mm height for generator
  inspection with the rotor still installed.
\newblock {\em IEEE Transactions on Industrial Electronics}, 58(12):5296--5303,
  Dec 2011.

\bibitem{Ratsamee_SSRR2016}
P.~Ratsamee, P.~Kriengkomol, T.~Arai, K.~Kamiyama, Y.~Mae, K.~Kiyokawa,
  T.~Mashita, Y.~Uranishi, and H.~Takemura.
\newblock A hybrid flying and walking robot for steel bridge inspection.
\newblock In {\em 2016 IEEE International Symposium on Safety, Security, and
  Rescue Robotics (SSRR)}, pages 62--67, Oct 2016.

\bibitem{Wang_TRO2017}
H.~Wang and A.~Yamamoto.
\newblock Analyses and solutions for the buckling of thin and flexible
  electrostatic inchworm climbing robots.
\newblock {\em IEEE Transactions on Robotics}, 33(4):889--900, Aug 2017.

\bibitem{La_IJFR2017}
H.~M. La, N.~Gucunski, K.~Dana, and S.-H. Kee.
\newblock Development of an autonomous bridge deck inspection robotic system.
\newblock {\em Journal of Field Robotics}, 34(8):1489--1504, 2017.

\bibitem{Mazumdar}
A.~Mazumdar and H.~H. Asada.
\newblock Mag-foot: A steel bridge inspection robot.
\newblock In {\em Intelligent Robots and Systems, 2009. IROS 2009. IEEE/RSJ
  International Conference on}, pages 1691--1696, Oct 2009.

\bibitem{RWang}
R.~Wang and Y.~Kawamura.
\newblock A magnetic climbing robot for steel bridge inspection.
\newblock In {\em Intelligent Control and Automation (WCICA), 2014 11th World
  Congress on}, pages 3303--3308, June 2014.

\bibitem{Fabien_IJFR2009}
F.~Tâche, W.~Fischer, G.~Caprari, R.~Siegwart, R.~Moser, and F.~Mondada.
\newblock Magnebike: A magnetic wheeled robot with high mobility for inspecting
  complex-shaped structures.
\newblock {\em Journal of Field Robotics}, 26(5):453--476, 2009.

\bibitem{Pack}
R.T. Pack, Jr. Christopher, J.L., and K.~Kawamura.
\newblock A rubbertuator-based structure-climbing inspection robot.
\newblock In {\em Robotics and Automation, 1997. Proceedings., 1997 IEEE
  International Conference on}, volume~3, pages 1869--1874 vol.3, Apr 1997.

\bibitem{Abderrahim}
M.~Abderrahim, C.~Balaguer, A.~Gimenez, J.~M. Pastor, and V.~M. Padron.
\newblock Roma: a climbing robot for inspection operations.
\newblock In {\em Robotics and Automation, 1999. Proceedings. 1999 IEEE
  International Conference on}, volume~3, pages 2303--2308 vol.3, 1999.

\bibitem{Leibbrandt}
A.~Leibbrandt, G.~Caprari, U.~Angst, R.~Y. Siegwart, R.J. Flatt, and
  B.~Elsener.
\newblock Climbing robot for corrosion monitoring of reinforced concrete
  structures.
\newblock In {\em Applied Robotics for the Power Industry (CARPI), the 2nd
  Intern. Conf. on}, pages 10--15, Sept 2012.

\bibitem{Rodriguez}
H.~Leon-Rodriguez, S.~Hussain, and T.~Sattar.
\newblock A compact wall-climbing and surface adaptation robot for
  non-destructive testing.
\newblock In {\em Control, Automation and Systems (ICCAS), 2012 12th
  International Conference on}, pages 404--409, Oct 2012.

\bibitem{SanMillan}
A.~San-Millan.
\newblock Design of a teleoperated wall climbing robot for oil tank inspection.
\newblock In {\em Control and Automation (MED), 2015 23th Mediterranean
  Conference on}, pages 255--261, June 2015.

\bibitem{Shen_MA20005}
W.~Shen, J.~Gu, and Y.~Shen.
\newblock Proposed wall climbing robot with permanent magnetic tracks for
  inspecting oil tanks.
\newblock In {\em IEEE International Conference Mechatronics and Automation,
  2005}, volume~4, pages 2072--2077 Vol. 4, July 2005.

\bibitem{Tavakol_RAS2013}
M.~Tavakoli, C.~Viegas, L.~Marques, J.~N. Pires, and A.~T. de~Almeida.
\newblock Omniclimbers: Omni-directional magnetic wheeled climbing robots for
  inspection of ferromagnetic structures.
\newblock {\em Robotics and Autonomous Systems}, 61(9):997 -- 1007, 2013.

\bibitem{Eich_MED2015}
M.~Eich and T.~Vögele.
\newblock Design and control of a lightweight magnetic climbing robot for
  vessel inspection.
\newblock In {\em 2011 19th Mediterranean Conference on Control Automation
  (MED)}, pages 1200--1205, June 2011.

\bibitem{Gilles_IJFR2012}
G.~Caprari, A.~Breitenmoser, W.~Fischer, C.~Hurzeler, F.~Tache, R.~Siegwart,
  O.~Nguyen, R.~Moser, P.~Schoeneich, and F.~Mondada.
\newblock Highly compact robots for inspection of power plants.
\newblock {\em Journal of Field Robotics}, 29(1):47--68, 2012.

\bibitem{Cho_TMECH2013}
K.~H. Cho, H.~M. Kim, Y.~H. Jin, F.~Liu, H.~Moon, J.~C. Koo, and H.~R. Choi.
\newblock Inspection robot for hanger cable of suspension bridge: Mechanism
  design and analysis.
\newblock {\em IEEE/ASME Transactions on Mechatronics}, 18(6):1665--1674, Dec
  2013.

\bibitem{La1}
H.~M. La, R.~S. Lim, B.~B. Basily, N.~Gucunski, J.~Yi, A.~Maher, F.~A. Romero,
  and H.~Parvardeh.
\newblock Mechatronic systems design for an autonomous robotic system for
  high-efficiency bridge deck inspection and evaluation.
\newblock {\em Mechatronics, IEEE/ASME Transactions on}, 18(6):1655--1664, Dec
  2013.

\bibitem{La2}
H.~M. La, N.~Gucunski, S.-H. Kee, and L.V. Nguyen.
\newblock Data analysis and visualization for the bridge deck inspection and
  evaluation robotic system.
\newblock {\em Visualization in Engineering}, 3(1):1--16, 2015.

\bibitem{La3}
H.~M. La, N.~Gucunski, S.H. Kee, and L.V. Nguyen.
\newblock Visual and acoustic data analysis for the bridge deck inspection
  robotic system.
\newblock In {\em The 31st International Symposium on Automation and Robotics
  in Construction and Mining (ISARC)}, pages 50--57, July 2014.

\bibitem{Lim1}
R.~S Lim, H.~M. La, Z.~Shan, and W.~Sheng.
\newblock Developing a crack inspection robot for bridge maintenance.
\newblock In {\em Robotics and Automation (ICRA), 2011 IEEE Intern. Conf. on},
  pages 6288--6293, May 2011.

\bibitem{Lim2}
R.~S. Lim, H.~M. La, and W.~Sheng.
\newblock A robotic crack inspection and mapping system for bridge deck
  maintenance.
\newblock {\em IEEE Transactions on Automation Science and Engineering},
  11(2):367--378, 2014.

\bibitem{Gibb_IJFR2018}
S.~Gibb, H.~M. La, T.~Le, L.~Nguyen, R.~Schmid, and H.~Pham.
\newblock Nondestructive evaluation sensor fusion with autonomous robotic
  system for civil infrastructure inspection.
\newblock {\em Journal of Field Robotics}, 0(0), 2018.

\bibitem{Sutter_AC2018}
B.~Sutter, A.~Lelevé, M.~T. Pham, O.~Gouin, N.~Jupille, M.~Kuhn, P.~Lulé,
  P.~Michaud, and P.~Rémy.
\newblock A semi-autonomous mobile robot for bridge inspection.
\newblock {\em Automation in Construction}, 91:111 -- 119, 2018.

\bibitem{Zhu_TMECH2012}
D.~Zhu, J.~Guo, C.~Cho, Y.~Wang, and K.~Lee.
\newblock Wireless mobile sensor network for the system identification of a
  space frame bridge.
\newblock {\em IEEE/ASME Transactions on Mechatronics}, 17(3):499--507, June
  2012.

\bibitem{Lee_RAS2013}
G.~Lee, G.~Wu, J.~Kim, and T.~Seo.
\newblock High-payload climbing and transitioning by compliant locomotion with
  magnetic adhesion.
\newblock {\em Robotics and Autonomous Systems}, 60(10):1308 -- 1316, 2012.

\bibitem{Seo_TMECH2013}
T.~Seo and M.~Sitti.
\newblock Tank-like module-based climbing robot using passive compliant joints.
\newblock {\em IEEE/ASME Transactions on Mechatronics}, 18(1):397--408, Feb
  2013.

\bibitem{Wang_ICA2014}
R.~Wang and Y.~Kawamura.
\newblock A magnetic climbing robot for steel bridge inspection.
\newblock In {\em Proceeding of the 11th World Congress on Intelligent Control
  and Automation}, pages 3303--3308, June 2014.

\bibitem{Guo_ICRA14}
J.~Guo, W.~Liu, and K.~M. Lee.
\newblock Design of flexonic mobile node using 3d compliant beam for smooth
  manipulation and structural obstacle avoidance.
\newblock In {\em 2014 IEEE International Conference on Robotics and Automation
  (ICRA)}, pages 5127--5132, May 2014.

\bibitem{Kamdar2015}
S.~Kamdar.
\newblock Design and manufacturing of a mecanum sheel for the magnetic climbing
  robot.
\newblock {\em Master Thesis, Embry-Riddle Aeronautical University}, May 2015.

\bibitem{Ward2015ClimbingRF}
P.~Ward, P.~Manamperi, P.~R. Brooks, P.~Mann, W.~Kaluarachchi, L.~Matkovic,
  G.~Paul, C.~H. Yang, P.~Quin, D.~Pagano, D.~Liu, K.~Waldron, and
  G.~Dissanayake.
\newblock Climbing robot for steel bridge inspection: Design challenges.
\newblock In {\em Austroads Publications Online, ARRB Group}, 2015.

\bibitem{PL_Allerton2016}
N.~H. Pham and H.~M. La.
\newblock Design and implementation of an autonomous robot for steel bridge
  inspection.
\newblock In {\em The 54th Annual Allerton Conference on Communication,
  Control, and Computing (Allerton)}, pages 556--562, Sept 2016.

\bibitem{Wang_IJIR2016}
R.~Wang and Y.~Kawamura.
\newblock Development of climbing robot for steel bridge inspection.
\newblock {\em Industrial Robot: An International Journal}, 43(4):429--447,
  2016.

\bibitem{PL_ISARC2016}
N.~H. Pham, H.~M. La, Q.~P. Ha, S.~N. Dang, A.~H. Vo, and Q.~H. Dinh.
\newblock Visual and 3d mapping for steel bridge inspection using a climbing
  robot.
\newblock In {\em The 33rd Intern. Symposium on Automation and Robotics in
  Construction and Mining (ISARC)}, pages 1--8, July 2016.

\bibitem{Takada_inventions2017}
Y.~Takada, S.~Ito, and N.~Imajo.
\newblock Development of a bridge inspection robot capable of traveling on
  splicing parts.
\newblock {\em Inventions}, 2(3), 2017.

\bibitem{La_Robotica_2018}
H.~M. La, T.~H. Dinh, N.~H. Pham, Q.~P. Ha, and A.~Q. Pham.
\newblock Automated robotic monitoring and inspection of steel structures and
  bridges.
\newblock {\em Robotica}, pages 1 -- 21, 2018.

\end{thebibliography}

\end{document}